\documentclass[10pt,twocolumn,letterpaper]{article}

\usepackage{wacv}              
\usepackage[accsupp]{axessibility}  

\definecolor{wacvblue}{rgb}{0.21,0.49,0.74}
\usepackage[pagebackref,breaklinks,colorlinks,allcolors=wacvblue]{hyperref}
\usepackage{graphicx}
\usepackage{amsmath}
\usepackage{multirow}
\usepackage{wrapfig}
\usepackage{algorithm}
\usepackage{algpseudocode}
\usepackage[table,dvipsnames]{xcolor}
\usepackage[capitalize]{cleveref}
\newcommand{\C}[1]{\mathcal{#1}}
\newcommand{\OP}[1]{\mathtt{#1}}
\newcommand{\algoname}{{\tt CalibBEV}}

\def\wacvPaperID{23} 
\def\confName{WACV}
\def\confYear{2026}

\title{CalibBEV: LiDAR-Camera Calibration via BEV Alignment}

\author{Filippo D'Addeo
\thanks{\scriptsize Equal contribution.} \hspace{0.1px}
\thanks{\scriptsize University of Bologna, Department of Industrial Engineering, Italy.}\\
{\tt\small filippo.daddeo2@unibo.it}
\and
Lorenzo Cipelli
\footnotemark[1] \hspace{0.1px}
\thanks{\scriptsize University of Parma, Department of Engineering and Architecture, Italy.}\\
{\tt\small lorenzo.cipelli@unipr.it}
\and
Adriano Cardace
\thanks{\scriptsize Stanford University, USA.}\\
{\tt\small cardace@stanford.edu}
\and
Emanuele Ghelfi
\thanks{\scriptsize VisLab srl, an Ambarella Inc. company, Italy.}\\
{\tt\small eghelfi@ambarella.com}
\and
Andrea Zinelli
\footnotemark[5]\\
{\tt\small azinelli@ambarella.com}
\and
Massimo Bertozzi
\footnotemark[3]\\
{\tt\small massimo.bertozzi@unipr.it}
}

\begin{document}
\maketitle
\begin{abstract}

We present \textbf{CalibBEV}, a novel Bird's Eye View (BEV) alignment approach for LiDAR-camera calibration. Our method unifies LiDAR and camera data into a shared 3D spatial representation, enabling accurate and robust cross-modal calibration.
CalibBEV extracts sensor-wise BEV features from each modality using domain-specific architectures and estimates the calibration matrix through a two-step alignment process. First, we perform an implicit alignment by regressing a coarse calibration matrix directly from the BEV features. To ease this alignment, we enforce semantic consistency between BEV representations across modalities using a contrastive loss inspired by CLIP, guiding both networks toward a unified feature space.
In the second step, we leverage our BEV formulation to explicitly align the features of one modality with the other, refining the initial coarse estimate into a final, more accurate calibration matrix.
CalibBEV significantly outperforms prior point-to-pixel matching methods, achieving state-of-the-art calibration accuracy. On the KITTI and nuScenes benchmarks, our method reduces the Relative Rotation Error (RRE) by 51\% and 68\%, and the Relative Translation Error (RTE) by 80\% and 91\%, respectively, compared to previous methods.

\end{abstract}
    
\section{Introduction}
\label{sec:intro}
Image-to-Point Cloud registration is crucial for applications in autonomous driving, 3D computer vision, augmented/virtual reality, and various robotics domains. The goal is to estimate the rigid transformation, including translation and rotation, that accurately aligns the projection of a LiDAR point cloud with a reference image.

Early works \cite{li2021deepi2p, ren2022corri2p} focus on extracting 2D and 3D features from RGB images and point clouds to establish 2D-3D correspondences. However, these methods often struggle with matching image and point cloud features, as different modalities and architectures do not inherently produce a shared feature space for seamless feature matching. To address this challenge, several attempts have been made:~\cite{Zhou2023VP2P} introduces a new voxel-based branch to learn more similar representations,~\cite{graphi2p} uses virtual point cloud to match corresponding representations of points, and~\cite{iclm} introduces correspondence queries to extract keypoints.

\begin{figure}[!tb]
    \centering
    \includegraphics[width=0.83\linewidth]{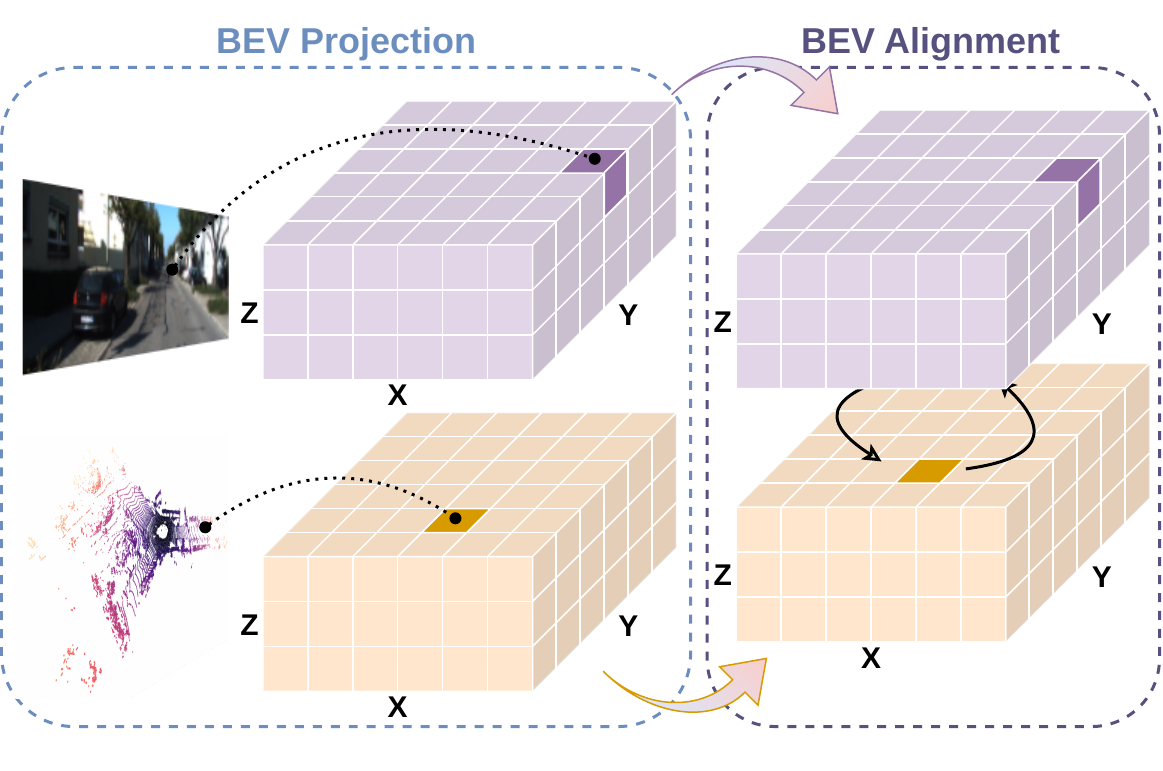}
    \caption{\textbf{BEV alignment.} Domain-specific features are computed independently and projected into 3D BEV representations. Then, our BEV alignment module estimates the calibration matrix between the two sensors.}
    \label{fig:teaser}
\end{figure}

In this work, we address the problem of Image-to-Point Cloud registration from a novel perspective, formulating it as a \textit{Bird's Eye View (BEV) alignment} task inspired by object detection literature \cite{centerpoint, simplebev, liftsplatshoot, bevformer, bevformer2, pointpillars}. As illustrated in \cref{fig:teaser}, we aim to generate two BEV feature representations of the same scene, one from a sparse LiDAR point cloud and the other from an RGB image, while leveraging modality-specific architectures without imposing any architectural constraints. 
Although the two BEV representations are derived from different modalities, they capture semantically similar scene structures, and we can frame Image-to-Point Cloud registration as the problem of aligning these BEV features.
The BEV alignment task is achieved with a two-step algorithm, referred to as implicit and explicit alignment.
First, in the implicit alignment, we consider the height dimension of the two BEVs as the channels of two top views, and feed them to a CNN-based decoder which is trained together with a rotation and a translation heads to predict a coarse rigid transformation.
Noticeably, we do not explicitly align the two BEVs. Instead, we let the decoder implicitly learn the registration parameters from the mis-aligned BEV feature maps by supervising the decoder and the prediction heads with the ground-truth rigid transformations. 
We argue that this is feasible due to the large receptive field of CNNs, which enables a global understanding of the BEV feature maps and allows the network to reason about the geometry of the two views for accurate calibration.

Although this strategy seems to be already effective as proven by our experiments reported in \cref{subsec:ablation}, a key improvement can be done by reasoning on the modality gap: the two BEV representations are generated by different networks and thus reside in distinct feature spaces. Simply feeding them into the same decoder results in suboptimal performance, as the decoder must simultaneously handle both alignment and feature discrepancies.
To address this, inspired by recent multi-modal alignment techniques~\cite{clip}, we adopt a CLIP loss that guides point features to be similar to their corresponding re-projected pixel features and vice-versa, leveraging known point-to-pixel correspondences during training.
This encourages modality-specific networks to learn a unified feature space representation. As a result, the BEV feature maps become more compatible, simplifying the Implicit Alignment task for the decoder and improving overall registration performance.

In the second step, we leverage our BEV formulation to perform an explicit alignment. Specifically, we use the coarse transformation matrix from the previous step to warp the LiDAR BEV features, reducing the misalignment with respect to the RGB BEV features. The aligned feature maps are then fed into a final decoder, which refines the calibration estimate to produce a final registration matrix.

Finally, unlike previous methods \cite{Zhou2023VP2P,curri2p,iclm,graphi2p}, we take one final step further by easily extending our model to leverage a multi-camera setup, in order to have a BEV representation with richer and denser RGB information.

To summarize, our contributions are:
\begin{itemize}
    \item We propose \algoname{}, a framework that for the first time frames the problem of Image-to-Point Cloud registration as a BEV alignment problem.
    \item Leveraging our unique BEV formulation, we propose a two-step alignment algorithm, where the first step performs an implicit alignment to estimate a coarse calibration, which is exploited in the explicit alignment step to obtain the final and refined calibration.
    \item We demonstrate that \algoname{} is easily extendable to support multi-camera setups, leading to better registration performances.
    \item \algoname{} achieves state-of-the-art-performance in both KITTI and nuScenes, surpassing previous methods by a significant margin while being more robust.
\end{itemize}
\begin{figure*}[!tb]
    \centering
    \includegraphics[trim={0 0 1.7cm 0},clip,width=0.95\textwidth]{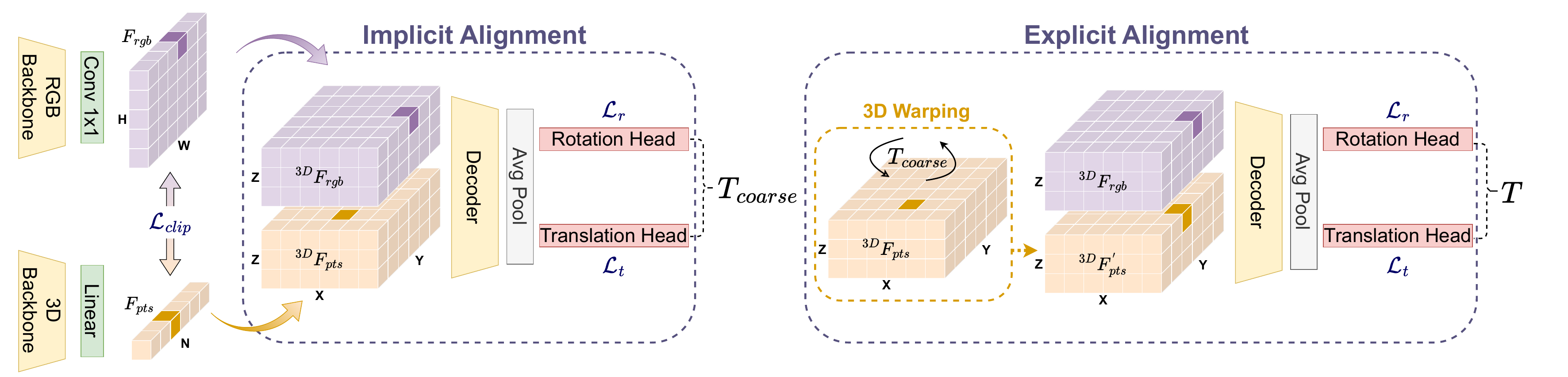}
    \caption{\textbf{Overview of our framework}. We process RGB images and point clouds independently, and adopt a CLIP loss between point and pixels features to encourage the networks to learn similar representations for corresponding pixel-point pairs. The extracted features are then lifted into two BEV representations, from which a decoder and prediction heads directly regress an initial coarse calibration matrix $T_{coarse}$ (\textbf{Implicit Alignment}).
    In the \textbf{Explicit Alignment} step, all previous network components are frozen, and the coarse calibration matrix is applied to warp the LiDAR BEV features toward the RGB BEV reference using a feature warping operator. Both the decoder and prediction heads are initialized from the previous step, and fine-tuned on the aligned features to produce the final calibration matrix $T$.
    }
    \label{fig:architecture}
\end{figure*}

\section{Related Work}
\label{sec:relatedworks}
\subsection{Image-to-Point Cloud Registration}

The LiDAR-Camera extrinsic calibration problem has been extensively studied, with early solutions relying on classical target-based and target-less methods. Target-based approaches \cite{cheng20233dradarcameracocalibration, 1334537, s110908992} use predefined objects and optimization techniques to align sensor perceptions. In contrast, target-less methods \cite{Levinson2013AutomaticOC, pandey} exploit environmental features or statistical models, resulting in more versatile approaches. Moreover, differently from the target-based, the target-less approaches can also be employed in online applications. Our approach falls into the target-less category. Latest state-of-the-art correspondence-based, target-less methods \cite{livox,directvisual,lcec} achieve remarkable registration performance. However, they typically involve heavy pre-processing, numerous frames accumulation, and accurate parameter fine tuning for each scene, resulting unsuitable for driving scenario, where scenes might drastically change in few seconds and fast inference is required. In contrast, \algoname{} doesn't require any particular pre-processing or accumulation technique.

\subsection{Projection-based Methods}
Projection-based methods estimate sensor calibration by projecting the LiDAR point cloud onto the image plane, generating a sparse depth map, and training a neural network to regress calibration parameters. RegNet~\cite{schneider2017regnetmultimodalsensorregistration} employs two separate backbones to extract features from the RGB image and sparse depth map, which are then fused for calibration estimation.
Similarly, CalibNet \cite{calibnet} enforces geometric consistency by ensuring that the projected sparse depth map, transformed with predicted extrinsics, aligns with the ground-truth sparse depth.
Building on these foundations, later methods introduced more advanced architectures: LCCNet \cite{lccnet} incorporates a feature matching layer to correlate RGB and depth features, while CalibFormer \cite{calibformer} leverages attention mechanisms for feature fusion.
Despite their effectiveness, projection-based methods have a key limitation: when there are no correspondences between the sparse depth projection and RGB, registration becomes impossible. In contrast, our approach directly processes RGB images and point clouds, accommodating uncalibrated sensors with rotations up to $\pm 360^\circ$.

\subsection{Point-based Methods}
Unlike projection-based methods, point-based algorithms operate directly in the 3D domain, processing point clouds instead of 2D sparse depth maps.
DeepI2P \cite{li2021deepi2p} is one of the first methods to leverage 3D architectures \cite{qi2017pointnetdeephierarchicalfeature,li2018sonetselforganizingnetworkpoint} to directly extract features from the point cloud. In its first stage, DeepI2P classifies whether each point lies within the camera frustum, or not. The second stage addresses the inverse camera projection problem by coarsely classifying points into corresponding image patches. Finally, a RANSAC-based PnP algorithm is applied to the grid classification output.
Similarly, CorrI2P \cite{ren2022corri2p} proposes an early classification stage to detect the overlapping region between the RGB and the point cloud while introducing cross-attention into the framework.
In the second stage, the camera pose can be obtained by applying EPnP \cite{epnp} within RANSAC on the dense correspondences between 2D and 3D features.
These methods often struggle with matching image and point cloud features, as different modalities and architectures (CNNs for images and MLPs for point clouds) do not inherently produce a shared feature space for seamless feature matching. To address this challenge, VP2P-Match~\cite{Zhou2023VP2P} introduces voxel-based representations into the framework exploiting an additional backbone. 
The voxel branch captures spatially local patterns, akin to the image branch, ensuring that pixels and voxels share similar feature structures. Finally, a differentiable probabilistic PnP solver \cite{chen2022epropnpgeneralizedendtoendprobabilistic} estimates the camera pose, making the network fully end-to-end differentiable.
Later, CurrI2P \cite{curri2p} improves both VP2P-Match and CorrI2P through the curriculum learning technique. 
However, these methods tend to be slow due to iterative nature of the outlier filtering methods such as RANSAC and the use of differentiable solvers, 
while also heavily relying on the classification of which points fall into the camera frustum.
For this reason, GraphI2P \cite{graphi2p} focuses on the feature consistency by estimating the camera pose given the LiDAR point cloud and a virtual cloud, obtained via depth estimation~\cite{zoedepth}, and by leveraging a point sampling on an unified spherical representation to build consistent features patterns, facilitating both the feature fusion and the graph-based correspondence selection phase.
Finally, ICLM \cite{iclm} introduces a set of learnable correspondence queries specialized as 2D and 3D keypoints detectors for estimating the final pose.
Similarly to previous approaches, we operate in a setting where features from the 2D and 3D domains are extracted from two independent networks and we adopt the same benchmark of point-based methods, consisting of a random vertical axis rotation in the range $\pm$360$^\circ$ and a random forward-backward and left-right axis translation in the range $\pm$10m. 

\section{Method}
\label{sec:method}

\subsection{Background}

As illustrated in~\cref{fig:architecture} (left), our network takes as input both an image $I \in \mathbb{R}^{H \times W \times 3}$ and a point cloud $P \in \mathbb{R}^{N \times 3}$ collected by an RGB camera and a LiDAR, respectively, where $H$ and $W$ are the height and the width of the images and $N$ is the number of points in the cloud.
Following previous works \cite{simplebev, Zhou2023VP2P, iclm}, we adopt two state-of-the-art 2D and 3D branches to independently encode the input modalities, opting for this approach over more complex multi-modal fusion architectures to achieve a lighter and faster network.
Building on \cite{simplebev}, we first encode images with a ResNet-50~\cite{resnet} and then we bilinearly upsample the features C4 of dimensions $H/16 \times W/16 \times C_4$, to $H/8 \times W/8 \times C_4$. These upsampled features are concatenated with the ResNet features C3 and finally encoded with a two-layer convolutional neural network (CNN), with instance normalization and ReLU activations, building the final 2D feature $F_{rgb}$ of size $H/8 \times W/8 \times C_{rgb}$.
In parallel, as in~\cite{Zhou2023VP2P}, we encode the point cloud using a Point Transformer~\cite{pointtransformer}, which consists of a downsampling and an upsampling module. Specifically, the downsampling applies a sequence of set abstraction layers~\cite{qi2017pointnet++} with a self-attention mechanism in order to obtain a global feature with dimensions $1 \times C_{pts}^{'}$, while the upsampling employs a set of feature propagation modules~\cite{qi2017pointnet++} computing the 3D point-wise features $F_{pts} \in \mathbb{R}^{N \times C_{pts}}$.
We refer to the supplementary material for a detailed graphical representation of both the backbones.
Subsequently, to ensure the same features dimensionality $C$ for both the RGB and point features, we encode the $F_{rgb}$ and $F_{pts}$ with a $1 \times 1$ convolutional layer and a single linear layer, respectively.
Then, we lift and project the 2D RGB and the 3D point cloud features into two 3D Bird's Eye View (BEV) spaces, and an Implicit Alignment module is introduced to estimate a coarse calibration matrix $T_{coarse}$ (see~\cref{subsec:implicit}) directly from the mis-registered BEVs features.
Finally, the coarse transformation is employed to explicitly align the BEV features of the 3D branch with the one from the RGB, and predict the final registration matrix (see~\cref{subsec:explicit}).

\subsection{Implicit Alignment}
\label{subsec:implicit}

The majority of the existing approaches handle the Image-to-Point Cloud registration as a classification task followed by a correspondence-matching phase, which enables estimating the mis-registration pose between the camera and the 3D sensor. In order to have a flexible architecture, we choose to formulate the Image-to-Point Cloud registration as a BEV alignment task. Indeed, we aim to obtain two BEV representations of the same scene: one from the RGB image and the other from the sparse LiDAR point cloud. For this reason, inspired by object detection literature~\cite{simplebev}, we first define a volume of 3D coordinates $G \in \mathbb{R}^{X \times Y \times Z \times 3}$, and  a zero-initialized 3D space $^{3D}F_{rgb} \in \mathbb{R}^{X \times Y \times Z \times C}$, representing the 3D coordinates and 3D feature space obtained from images surrounding the vehicle.
We employ a East-Down-North (EDN) coordinate system, centering the volume with the camera frame origin. Then, in order to sample and lift the 2D features from $F_{rgb}$ to $^{3D}F_{rgb}$, the 3D coordinates $G$ are projected from the camera EDN frame $\{ c \}$ onto the image plane with the intrinsics matrix $K$ through~\cref{eq:img_proj} to get the corresponding 2D coordinates:
\begin{equation}
\label{eq:img_proj}
  \begin{bmatrix} u & v & d \end{bmatrix}^T = K \cdot g, \ \ \forall g = \begin{bmatrix} {}^cx & ^cy & ^cz\end{bmatrix}^T \in G ,
\end{equation}

and we sample features given the 2D coordinates to fill the corresponding cell in $^{3D}F_{rgb}$.

Similarly, we define a zero-initialized 3D space $^{3D}F_{pts} \in \mathbb{R}^{X \times Y \times Z \times C}$. The mis-aligned point cloud is then projected onto this three-dimensional space and the features $F_{pts}$ are scattered into $^{3D}F_{pts}$ accordingly to the corresponding projection cells. Afterward, we flatten the $Y$ and $C$ axes both in $^{3D}F_{rgb}$ and $^{3D}F_{pts}$, obtaining BEV feature maps with dimensions $X \times Z \times \left( Y \cdot C \right)$. 
We then concatenate the two BEVs along the channel dimension, and feed them to a single $3 \times 3$ convolutional layer $\varphi(\cdot)$ followed by instance normalization and a non-linearity for fusing the two modalities:

\begin{equation}
\label{eq:fusion}
    {^{3D}F_{fused}} = \varphi \left(^{3D}F_{rgb} \oplus ^{3D}F_{pts} \right) ,
\end{equation}

with $\oplus$ being the concatenation. Finally, the fused features are processed by a convolutional decoder which, thanks to its large receptive field, can reason globally about the geometry and misalignment between the two BEVs, producing high-level features that benefit the rotation and translation heads. It is important to note that although the two BEVs are not explicitly aligned, we empirically found that the decoder is sufficiently powerful to extract the necessary features to solve the calibration task, as demonstrated by our experiments.
To this end, we adopt a ResNet-18~\cite{resnet} architecture as the decoder, where the average-pooled C5 features serve as input to the translation and rotation prediction heads. Both heads are implemented as two-layer MLPs with ReLU non-linearity.
We refer to the supplementary material for further implementation details.
For the translation component, we directly predict the translation vector $t = \begin{bmatrix} x & y & z \end{bmatrix}^T$. Conversely, for the rotation component, we estimate the sine and cosine values of the rotation angles along each axis. From these predictions, we construct the corresponding rotation matrices $R_z(\psi)$, $R_y(\phi)$, and $R_x(\theta)$, which are then combined to decode the coarse calibration matrix as:

\begin{equation}
    T_{coarse} = \begin{bmatrix} R_z(\psi) \cdot R_y(\phi) \cdot R_x(\theta) & t\\
                                                      0 & 1\end{bmatrix} .
\end{equation}

As discussed in \cref{sec:intro} and demonstrated by our experiments (see \cref{sec:experiments}), the decoder effectively aligns the concatenated feature maps ${^{3D}F_{fused}}$, yielding accurate calibration matrix estimates.
Nonetheless, we propose to further boost performance by adopting a CLIP-based~\cite{clip} loss function that encourages $^{3D}F_{rgb}$ and $^{3D}F_{pts}$ to reside in a unified feature space. 
In particular, we aim at guiding the two backbones to learn similar features for corresponding pixel-point pairs.
This alignment facilitates the decoder's task, allowing it to focus solely on spatial alignment rather than compensating for modality-specific feature discrepancies.
During training, feature maps from the 2D and 3D branches are first normalized row-wise obtaining the normalized features $\tilde{F}$ as follows:
\begin{equation}
\tilde{F} = D^{-1} \cdot F,  \ \ F \in \{F_{rgb}, F_{pts}\} ,
\end{equation}

with $D$ being a square diagonal matrix with $ D_{ii} = \| f_i \|_2 $ and $i$ indicating the i-th row. Then, the similarity matrix $S\in \mathbb{R}^{(H/8 \times W/8) \times N}$ between $\tilde{F}_{rgb}$ and $\tilde{F}_{pts}$ is computed as follows:

\begin{equation} S = e^{\gamma} \left( \tilde{F}_{rgb} \cdot \tilde{F}^T_{pts} \right), \end{equation}

where $\gamma$ is a learnable temperature parameter.

In order to build the target one-hot 2D-3D correspondences matrix $\hat{S} \in \mathbb{R}^{(H/8 \times W/8) \times N}$, we rely on the known ground truth registration matrix $\begin{bmatrix}\hat{R} & \hat{t}\end{bmatrix}$ at training time to correctly transform the point cloud into camera coordinates and project it into the image plane to find the corresponding pixel-point pairs:

\begin{equation}
\label{eq:pts_to_img}
  \begin{bmatrix} u & v & d \end{bmatrix}^T = K \left( \hat{R} \ p + \hat{t} \right) \ \ \forall p \in P .
\end{equation}

Following CLIP~\cite{clip}, we compute the average symmetric cross-entropy loss in both directions:
\begin{equation}
\label{eq:clip_loss}
    \C{L}_{clip} = \frac{\C{L}_{\OP{CE}(\sigma(S), \hat{S})} + \C{L}_{\OP{CE}(\sigma(S^T), \hat{S}^T)}}{2} ,
\end{equation}

with $\sigma$ being the Softmax operator.

At the same time, we supervise the rotation and translation heads. To this end, we employ an L1 loss $\C{L}_{r}$ for supervising the sine and the cosine rotation values along each axis, while we minimize the L2 norm difference $\C{L}_{t}$ between the predicted and the ground truth translation vectors to compute the translation error. 
The overall optimization function is:

\begin{equation}
\label{eq:loss}
    \C{L} = \alpha \ \C{L}_{clip} + \C{L}_{r} + \C{L}_{t} ,
\end{equation}

where $\alpha$ is the CLIP loss weighting factor.

\subsection{Explicit Alignment}
\label{subsec:explicit}
Peculiar to our BEV formulation, we can exploit the coarse transformation matrix from the Implicit Alignment step to explicitly align $^{3D}F_{pts}$ with $^{3D}F_{rgb}$, as depicted in \cref{fig:architecture} (right), 
and provide partially aligned feature maps to the decoder and the prediction heads, leading to a better estimate of the calibration matrix. 
Specifically, in the second step, we freeze the Implicit Alignment module and we employ $T_{coarse}$, to pre-align the 3D points features, $^{3D}F_{pts}$, with the RGB features $^{3D}F_{rgb}$ obtaining $^{3D}F^{'}_{pts}$. 
We do this by applying a differentiable sampling mechanism~\cite{stn} in which each feature vector at location $g$ in $^{3D}F^{'}_{pts}$ is sampled from a position $\bar{g} = T_{coarse}(g)$ in $^{3D}F_{pts}$ in the following way:

\begin{equation}
\label{eq:warp_refine}
    ^{3D}F^{'}_{pts}(g) = \sum_{g_i \in \mathcal{N}(\bar{g})} w_{g_i} {^{3D}F_{pts}(g_i)} ,
\end{equation}

where $w_{g_i}$ and $\mathcal{N}$ are the bi-linear kernel weights obtained from $T_{coarse}$ and the set of neighboring pixels, respectively. Hence, \cref{eq:warp_refine} can be considered as a backward warping, where the new features in $^{3D}F^{'}_{pts}$ are obtained by warping $^{3D}F_{pts}$ according to $T_{coarse}$.
Finally, $^{3D}F_{rgb}$ and $^{3D}F^{'}_{pts}$ are fused following \cref{eq:fusion}, and a new decoder with the corresponding prediction heads is employed to predict the fine calibration matrix $T_{fine}$.

The final transformation matrix $T$ is obtained by combining the coarse and fine calibration matrix:

\begin{equation}
\label{eq:final_T}
T = T_{fine} \cdot T_{coarse} .
\end{equation}

Notably, both the decoder and prediction heads of the Explicit Alignment module adopt the same architecture as in the previous stage. We initialize their weights from the previous step for faster convergence, employing the same loss functions to supervise the final transformation matrix $T$ as before.
\begin{table*}[!tb]\small
  \setlength{\abovecaptionskip}{2mm}
  \centering
  \resizebox{0.8\textwidth}{!}{
  \begin{tabular}{l|c|c|c|c|c|c}
    \toprule
    \multirow{1}*{} & \multicolumn{3}{c|}{\cellcolor{blue!25}KITTI} & \multicolumn{3}{c}{\cellcolor{YellowOrange}nuScenes}\\
    \cline{2-7}
    
    Method  &RTE(m)$\downarrow$ &RRE($^\circ$)$\downarrow$ &Acc.$\uparrow$ &RTE(m)$\downarrow$ &RRE($^\circ$)$\downarrow$ &Acc.$\uparrow$\\
    \midrule
    Grid Cls. + PnP \cite{li2021deepi2p} & 3.64 $\pm$ 3.46 & 19.19 $\pm$ 28.96 & 11.22 & 3.02 $\pm$ 2.40 & 12.66 $\pm$ 21.01 & 2.45 \\
    DeepI2P (3D) \cite{li2021deepi2p} & 4.06 $\pm$ 3.54 & 24.73 $\pm$ 31.69 & 3.77 &  2.88 $\pm$ 2.12 & 20.65 $\pm$ 12.24 & 2.26 \\
    DeepI2P(2D) \cite{li2021deepi2p} & 3.59 $\pm$ 3.21 & 11.66 $\pm$ 18.16 & 25.95 & 2.78 $\pm$ 1.99 & 4.80 $\pm$ 6.21 & 38.10 \\
    CorrI2P \cite{ren2022corri2p} & 3.78 $\pm$ 65.16 & 5.89 $\pm$ 20.34 & 72.42 & 3.04 $\pm$ 60.76 & 3.73 $\pm$ 9.03 & 49.00 \\
    CurrI2P (CorrI2P) \cite{curri2p} & 1.55 $\pm$ 7.99 & 3.61 $\pm$ 10.88 & - & 2.16 $\pm$ 4.20 & 2.98 $\pm$ 5.35 & - \\
    VP2P-Match \cite{Zhou2023VP2P} & 0.75 $\pm$ 1.13 & 3.29 $\pm$ 7.99 & 83.04 & 0.89 $\pm$ 1.44 & 2.15 $\pm$ 7.03 & 88.33 \\
    CurrI2P (VP2P-Match) \cite{curri2p} & 0.53 $\pm$ 1.00 & 2.11 $\pm$ 9.48 & - & 1.04 $\pm$ 1.64 & 2.67 $\pm$ 8.61 & - \\
    RelaI2P \cite{relai2p} & 0.72 $\pm$ 1.45 & 2.92 $\pm$ 6.67 & 85.60 & - & - & - \\
    ICLM \cite{iclm} & \underline{0.20} $\pm$ 0.21 & \underline{1.24} $\pm$ 2.34 & 97.49 & 0.63 $\pm$ 0.44 & 2.13 $\pm$ 3.75 & 90.94 \\
    GraphI2P \cite{graphi2p} & 0.32 $\pm$ 0.81 & 1.65 $\pm$ 1.32 & \underline{99.61} & \underline{0.49} $\pm$ 1.22 & \underline{1.73} $\pm$ 1.63 & \underline{99.48} \\
    \midrule
    \algoname{} (ours) & \textbf{0.04 $\pm$ 0.10} & \textbf{0.61 $\pm$ 0.52} & \textbf{99.96} & \textbf{0.04 $\pm$ 0.08} & \textbf{0.54 $\pm$ 0.45} & \textbf{99.98} \\
    \bottomrule
  \end{tabular}}
  \caption{Registration results on the KITTI odometry and nuScenes datasets. Lower is better for both RTE and RRE, while higher is better for accuracy. Values are from original papers and '-' due to missing results and training code.}
  \label{table:sota}
\end{table*}

\section{Experiments}
\label{sec:experiments}

\subsection{Datasets and Metrics}
\label{subsec:dataset}

We use the public KITTI~\cite{kitti} and nuScenes~\cite{nuscenes} datasets for training and evaluating our network. Regarding the KITTI dataset, we use the odometry benchmark which consists of 22 stereo sequences containing the images collected by two RGB stereo cameras facing forward the vehicle and the point cloud acquired by a LiDAR mounted on top of the vehicle. Following previous works, we use the sequences from 00 to 08 for training and the sequences 09 and 10 for testing. In addition, we employ the images acquired by both the RGB stereo cameras both for training and testing. 
The nuScenes dataset consists of 1000 different scenes containing the images collected by 6 different cameras returning a $360^\circ$ Field Of View (FOV) of the scene around the vehicle and the point clouds acquired by 5 radars and a LiDAR. Following previous works, we use the official 700 sequences in the train split and the 150 scenes in the evaluation split for training, while we use the remaining 150 scenes of the test split for testing. We use only the images acquired by the front camera and the point cloud acquired by the LiDAR for both training and testing.

As evaluation metrics, we follow previous works~\cite{Zhou2023VP2P,iclm,graphi2p}, and we evaluate the registration performance with the average Relative Translation Error (RTE) and the average Relative Rotation Error (RRE)~\cite{rre_rte}. Moreover, for a complete comparison with other works, we also report the registration accuracy (Acc.) \cite{Zhou2023VP2P} which corresponds to the proportions of registrations achieving RTE $< 2$m and RRE $< 5^\circ$, even if it does not properly asses the fine-grained quality of predicted registrations. Finally, we report the RTE and RRE standard deviations, computed by aggregating all the RTE and RRE values for each test sample.

\subsection{Implementation Details}
\label{subsec:implementation}

As done in~\cite{curri2p,iclm,graphi2p}, for the KITTI dataset, we set the input RGB images dimensions to be $160 \times 512$. These are obtained by removing the top-50 rows, applying a $0.5$ downsample factor, and randomly cropping the resulting images to match the desired dimensions. Afterward, we downsample the point cloud size to $N = 40960$.
Similarly, for nuScenes dataset, we set images to be of size $160 \times 320$. These are obtained by removing the top-100 rows, applying a $0.2$ downsample factor, and randomly cropping the resulting images to the desired dimensions. In regard to the 3D branch, we set the point cloud dimension $N = 40960$ by accumulating from the past frames, since a single LiDAR sweep in nuScenes contains roughly 36k points. However, this is not a required step for \algoname{}. Hence, we refer to the supplementary material for a further analysis.
In the alignment modules, we set X, Y, and Z to  200, 8 and 200, respectively, spanning a 3D metric space of $\pm25$m in both the forward-backward and left-right directions and of $\pm 5$m in up-bottom directions, with features dimension C to 128.

We set the batch size to 8 in all our experiments and train for 260k iterations using Adam~\cite{adam} as optimizer. We use $1 \times 10^{-4}$ as initial learning rate, which is decayed by a $0.5$ factor every 51k iterations. Moreover, following~\cite{clip}, we initialize the learnable temperature parameter $\gamma$ defined in~\cref{subsec:implicit} to $\log(1/0.07)$. Finally, we set the $\C{L}_{clip}$ weighting factor $\alpha$ to $0.5$.

As regards the Explicit Alignment module, after completely training the Implicit Alignment, both the new decoder and the prediction heads are initialized with the weights of their corresponding components from the previous step and both the backbones are frozen, as well as the Implicit Alignment decoder and prediction heads, leaving only the Explicit Alignment module parameters to be learnable. Hence, we train the Explicit Alignment module for an additional 120k iterations, setting the learning rate to $1 \times 10^{-5}$ for all the learnable parameters and decayed, once again, by a $0.5$ factor every 51k iterations. 
For all remaining hyper-parameters, we keep them as stated above.

\begin{figure}[t]
    \centering
    \includegraphics[width=\linewidth]{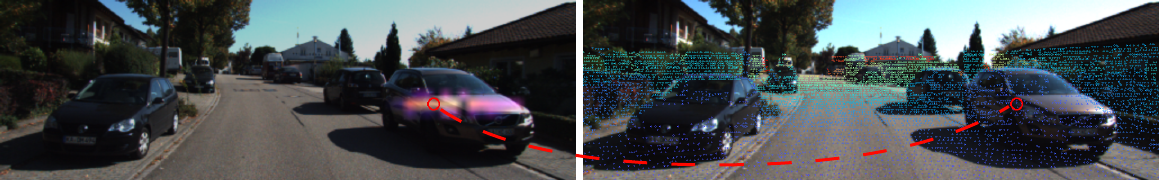}
    \caption{\textbf{2D-3D feature similarity heatmap.} For a given 3D point projected into the image plane (right), we highlight the most similar pixels in features space (left).}
    \label{fig:clip}
\end{figure}

\begin{figure*}[!tb]
    \centering
    \includegraphics[trim={0 7px 0cm 0},clip,width=\textwidth]{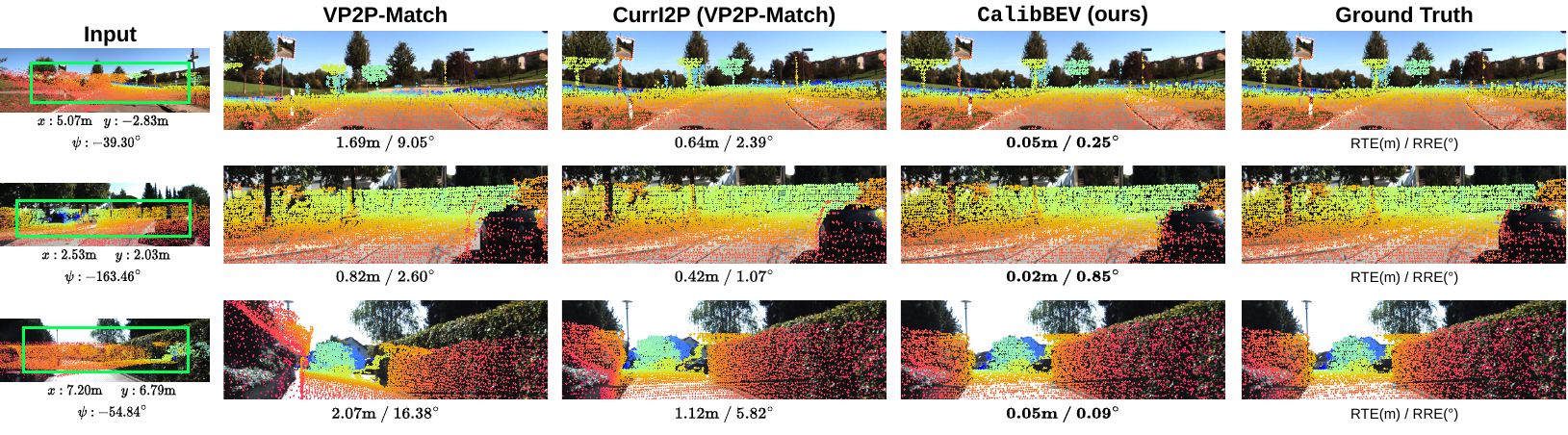}
    \caption{
    Qualitative comparison of Image-to-Point Cloud registration results on the KITTI dataset. From left to right: mis-aligned point cloud and image inputs, VP2P-Match calibration result, our model calibration result, and the ground truth camera-LiDAR alignment. 
    }
    \label{fig:qualitatives}
\end{figure*}

\subsection{Comparison with State-of-the-Art}
\label{subsec:sota}
In \cref{table:sota}, we compare the performance of \algoname{} with the latest state-of-the-art methods according to the standard benchmark, which consists in sampling a random rotation along the vertical axis in the range $\pm$360$^\circ$ and a random translation along the forward-backward and left-right directions in the range $\pm$10m. We also follow~\cite{Zhou2023VP2P,iclm,graphi2p} and do not remove the samples with large errors before computing the metrics, which would result in an unsuitable way to report the real performance. We refer to the supplementary material for this specific evaluation.
On the KITTI dataset, \algoname{} outperforms ICLM, which is to the best of our knowledge the current state-of-the-art model on the KITTI dataset, by 0.16m on the average RTE, by 0.63$^\circ$ on the average RRE and by 2.47\% on the registration accuracy. We also highlight \algoname{} is more robust compared to all previous works. Indeed, our proposal results in a reduction in the standard deviation of 0.11m and 1.82$^\circ$ on the RTE and RRE, respectively.
Similarly, on the nuScenes dataset, \algoname{} outperforms the current state-of-the-art model GraphI2P by 0.45m, 1.19$^\circ$, and 0.5\% on the RTE, RRE, and registration accuracy, respectively. Once again, \algoname{} demonstrates superior robustness compared to all prior works, as it is able to reduce the standard deviation by 0.45m and by 1.18$^\circ$ on the RTE and RRE, respectively.

\subsection{Ablation Studies}
\label{subsec:ablation}

In this section, we conduct ablation studies to highlight how different parameters and architectural choices reflect on the registration task. We report the effectiveness of each contribution of our work in~\cref{table:ablation_components}. Inspired by \cite{li2021deepi2p}, we test a baseline architecture just composed by the modality-specific backbones, recovering one single global feature from the RGB and one from the point cloud, directly regressing the calibration from their concatenation. This architecture suffers from a large error in both rotation and translation, caused by the lack of spatial information, serving only as a reference with respect to further methods.
Then, we ablate the effect of the Implicit Alignment module (second row of \cref{table:ablation_components}) to gain spatial information by leveraging two BEV representations from both the RGB image and the LiDAR point cloud. We highlight that our first contribution already achieves comparable results to the state-of-the-art model on KITTI, while outperforming other methods on nuScenes. This result endorses our choice to leverage the geometry of the BEV representation as core for the calibration estimation.

\begin{table}[tb!]
    \centering
    \scalebox{0.7}{
    \begin{tabular}{ccc|cc|cc}
    \toprule
        & & & \multicolumn{2}{c|}{\cellcolor{blue!25} KITTI} & \multicolumn{2}{c}{\cellcolor{YellowOrange}nuScenes}\\
        \cline{4-7}
        
        Implicit & $\mathcal{L}_{clip}$ & Explicit & RTE(m)$\downarrow$ &RRE($^\circ$)$\downarrow$ & RTE(m)$\downarrow$ &RRE($^\circ$)$\downarrow$\\
        \midrule
                   &            &            & 1.07 & 13.73 & 0.71 & 10.83\\
        \checkmark &            &            & 0.04 & 1.61 & 0.07 & 1.48 \\
        \checkmark & \checkmark &            & 0.05 & 1.17 & 0.04 & 1.16 \\
        \checkmark & \checkmark & \checkmark & \textbf{0.04} & \textbf{0.61} & \textbf{0.04}    & \textbf{0.54} \\
        \bottomrule
    \end{tabular}}
    \caption{Ablation studies of each component. Implicit: Implicit Alignment module, $\mathcal{L}_{clip}$: effect of CLIP loss between corresponding point-pixels pairs, Explicit: Explicit Alignment module.}
    \label{table:ablation_components}
\end{table}

Next, by adding a CLIP-like loss to encourage the similarity between RGB features and corresponding point features and vice versa, we are able to reduce the RRE by 0.44$^\circ$ and 0.32$^\circ$ on KITTI and nuScenes respectively (third row of \cref{table:ablation_components}). We argue that this result shows the effectiveness of a well-established loss function in enhancing the similarity between cross-domain features, leading to a positive impact on the \algoname{} performance by simplifying the decoder task. 
We further illustrate the effect of $\C{L}_{clip}$ in \cref{fig:clip}, where a selected 3D point (circled in red, bottom figure) is projected onto the image plane for visualization. The most similar pixels in feature space are highlighted on the image (top figure). As shown, for a 3D point belonging to the car on the right side, the highest-scoring pixels consistently lie within the car region, demonstrating that both modality-specific networks learn semantically consistent features for corresponding objects across modalities.

Lastly, we test our Explicit Alignment module, where we explicitly align the BEV geometries from the RGB and the point cloud as explained in \cref{subsec:explicit}. This contribution leads to an RRE reduction of 0.56$^\circ$ and 0.62$^\circ$ on KITTI and nuScenes, respectively (last row of \cref{table:ablation_components}), supporting our choice to exploit the realignment of features at spatial level through the BEV representation. \\

\begin{table}[tb!]
    \centering
    \scalebox{0.7}{
    \begin{tabular}{cc|ccc}
    \toprule
        & & \multicolumn{3}{c}{\cellcolor{blue!25}KITTI} \\
        \cline{3-5}
        
        Train & Test & RTE(m)$\downarrow$ & RRE($^\circ$)$\downarrow$ & Acc.$\uparrow$\\
        \midrule
        L+R & R & 0.04 $\pm$ 0.10 & 0.61 $\pm$ 0.52 & 99.96\\
        L   & R & 0.03 $\pm$ 0.02 & 0.72 $\pm$ 0.61 & 99.89\\
        \bottomrule
    \end{tabular}}
    \caption{Analysis on \algoname{} generalization ability when trained on left camera images and tested on right ones.}
    \label{table:ablation_leftright}
\end{table}

\begin{table}[!tb]
    \centering
    \scalebox{0.7}{
    \begin{tabular}{cc|ccc}
    \toprule
        & & \multicolumn{3}{c}{\cellcolor{YellowOrange}nuScenes}\\
        \cline{3-5}
        Method & Num. Cams. & RTE(m)$\downarrow$ & RRE($^o$)$\downarrow$ & Acc.$\uparrow$ \\
        \midrule
        \algoname{} (ours) & 1 & 0.04 $\pm$ 0.08 & 0.54 $\pm$ 0.45 & 99.98\\
        \algoname{} (ours) & 6 & \textbf{0.04 $\pm$ 0.03} & \textbf{0.28 $\pm$ 0.22} & \textbf{100.0}\\
        \bottomrule
    \end{tabular}}
    \caption{Multi-camera contribution analysis.}
    \label{table:ablation_multicam}
\end{table}

\noindent \textbf{Generalization Analysis.} In zero-shot generalization, different scenes or sensors inter-calibrations may weaken the registration performance if the model training leaned towards overfitting a particular sensor suite. On account of this problem, we decided to investigate the \algoname{} generalization ability when trained solely on the left camera images and evaluated on the right camera ones on the KITTI dataset. In~\cref{table:ablation_leftright}, we show how \algoname{} is comfortably able to generalize, even when the camera position changes between training and testing. Indeed, the registration performance doesn't get undermined when the model is trained solely on the left camera images and evaluated on the right camera ones, compared to the \algoname{} trained on both the cameras images and evaluated on the right ones.\\

\begin{table}[tb!]
    \centering
    \scalebox{0.7}{
    \begin{tabular}{c|ccc}
    \toprule
        & \multicolumn{3}{c}{\cellcolor{blue!25}KITTI} \\
        \cline{2-4}
        
        Method & RTE(m)$\downarrow$ & RRE($^\circ$)$\downarrow$ & Acc.$\uparrow$\\
        \midrule
        CorrI2P \cite{ren2022corri2p} & 0.96 $\pm$ 3.10 & 2.87 $\pm$ 4.58 & 85.76 \\
        \midrule
        Calibnet \cite{calibnet} & 5.88 $\pm$ 2.80 & 10.92 $\pm$ 6.09 & 3.03 \\
        LCCNet \cite{lccnet} & 0.40 $\pm$ 0.29 & 4.27 $\pm$ 3.70 & 75.75 \\
        \midrule
        \algoname{} (ours) & \textbf{0.10 $\pm$ 0.06} & \textbf{1.13 $\pm$ 0.62} & \textbf{99.96} \\
        \bottomrule
    \end{tabular}}
    \caption{6DoF registration analysis. For each axis sample a rotation between $\pm$20$^\circ$ and a translation between $\pm$1.5m.}
    \label{table:ablation_6dof}
\end{table}

\begin{table}[tb!]
    \centering
    \scalebox{0.7}{
    \begin{tabular}{ccc|ccc}
    \toprule
        & & & \multicolumn{3}{c}{\cellcolor{blue!25}KITTI} \\
        \cline{4-6}
        
        BEV & Range(m) & Cell(m) & RTE(m)$\downarrow$ & RRE($^\circ$)$\downarrow$ & Acc.$\uparrow$\\
        \midrule
        200 $\times$ 200 & $\pm$100 & 1.0 & 0.11 & 2.08 & 93.05\\
        200 $\times$ 200 & $\pm$75 & 0.75 & 0.10 & 1.82 & 96.61\\
        200 $\times$ 200 & $\pm$50 & 0.50 & 0.07 & 1.18 & 99.55\\
        200 $\times$ 200 & $\pm$25 & 0.25 & \textbf{0.04} & \textbf{0.61} & \textbf{99.96}\\
        400 $\times$ 400 & $\pm$50 & 0.25 & 0.06 & 0.78 & 99.76\\
        \bottomrule
    \end{tabular}}
    \caption{BEV hyperparameters analysis.}
    \label{table:ablation_bev}
\end{table}

\noindent \textbf{Multi-camera Analysis.} In~\cref{table:ablation_multicam}, we explore the contribution for multi-camera configurations. Indeed, real-world applications and modern datasets such as \cite{nuscenes,waymo,kitti360,argoverse2} are typically characterized by multi-camera setups, returning a 360$^\circ$ FOV of the scene surrounding the vehicle. Differently from previous works~\cite{Zhou2023VP2P,iclm,graphi2p}, \algoname{} is easily extendable to multi-camera configurations by simply adding few lines of modifications in the RGB to BEV projection. Specifically, the RGB backbone encodes each input image obtaining 2D features $F_{rgb}$ of size $NC \times H/8 \times W/8 \times C_{rgb}$, with NC being the number of cameras. Subsequently, for each camera, these features are sampled and lifted from $F_{rgb}$ to $^{3D}F_{rgb}$ as described in~\cref{subsec:implicit}. Finally, following~\cite{simplebev} we apply a weighted-sum for fusing the different volumes computed for each camera, taking into account for intersection between adjacent camera frustums by averaging the features projected in these regions. 
In this way, when dealing with multiple cameras, \algoname{} demonstrates better calibration performances, outperforming by 0.48\% in RRE the single camera setup, while being more robust in all the metrics, highlighting a 100\% registration accuracy.\\

\noindent \textbf{6DoF Registration Analysis.} Although this is not the standard benchmark for point-based methods, in \cref{table:ablation_6dof} we show \algoname{} registration performances in a Six Degrees of Freedom (6DoF) scenario, where variations for the three spatial coordinates and rotation angles are involved. Following \cite{lccnet}, for each axis we sample both a random rotation and translation in range $\pm 20^\circ$ and $\pm 1.5$m, respectively. \algoname{} outperforms the latest state-of-the-art point-based approach with publicly available training code CorrI2P \cite{ren2022corri2p} by 0.86m , 1.74$^\circ$, and 14.2\% on the RTE, RRE, registration accuracy, respectively, also demonstrating the \algoname{} ability to estimate the height by encoding it into the channels dimension.
For a fair and complete analysis, we also compare \algoname{} against Calibnet \cite{calibnet} and LCCNet \cite{lccnet}, two state-of-the-art projection-based approaches. We outperform Calibnet by 5.78m, 9.79$^\circ$, and 96.93\% on the RTE, RRE, registration accuracy, respectively, while we surpass LCCNet by 0.30m, 3.14$^\circ$, and 24.21\% on the RTE, RRE, registration accuracy, respectively. Once again, \algoname{} shows to be more robust compared to all previous works that have been tested.\\

\noindent \textbf{BEV Resolution Analysis.} In~\cref{table:ablation_bev} we compare different 3D grid resolutions. We found an optimal setup by setting the cell dimension to 0.25m along the forward-backward and left-right directions. Specifically, we limit the BEV dimensions to 400 $\times$ 400, since higher resolutions would sacrifice the inference time.\\

\noindent \textbf{Qualitative Analysis.} A qualitative analysis on the KITTI dataset is shown in~\cref{fig:qualitatives}. For visualization, we first apply the predicted alignment transformation matrix to the point cloud, which is then projected onto the image plane through the known camera intrinsics parameters. We compare \algoname{} (fourth column from left) against the state-of-the-art models CurrI2P and VP2P-Match, and the ground truth (last column from left), reporting the RTE and RRE for each prediction. Instead, in the first column, we show the mis-registered image and point cloud inputs. Notably, we highlight the \algoname{} ability to achieve better alignment performance compared to CurrI2P in complex scenarios, in which the network might struggle to extract distinctive visual features. Indeed, by leveraging BEV features, our model demonstrates to have a better understanding of the geometrical relationships in the surrounding environment.
Finally, we refer to the supplementary material for a further analysis on the nuScenes dataset and the failure cases.

\section{Conclusion}
\label{sec:conclusion}
In this paper we introduced \algoname{}, the first framework to cast the LiDAR-camera calibration problem as a BEV alignment task. Our key insight is to extract 3D BEV representations from both modalities and directly estimate the calibration matrix by aligning these representations through an Implicit Alignment module. Leveraging our unique BEV formulation, we further proposed an Explicit Alignment step, where the initial coarse estimate is used to warp the BEV features and refine the calibration matrix.
Extensive experiments demonstrate that our approach achieves state-of-the-art performance, surpassing previous methods by a significant margin in terms of accuracy and robustness.
Future works may include developing an end-to-end training pipeline to simultaneously optimize both the Implicit and Explicit Alignments, avoiding a two-step strategy. 
{
    \small
    \bibliographystyle{ieeenat_fullname}
    \bibliography{main}
}
\end{document}